\documentclass{ecai}
\usepackage{times}
\usepackage{graphicx}
\usepackage{latexsym}
\usepackage{cite}
\usepackage{amsmath,amssymb,amsfonts}
\usepackage{textcomp}
\usepackage{xcolor}
\usepackage{algorithm,algorithmic}
\usepackage{multirow}
\usepackage{booktabs}
\usepackage{array}
\usepackage{stfloats}
\usepackage{subfigure}
\usepackage{color}
\usepackage{caption2}
\usepackage{url}
\usepackage[numbers,sort&compress]{natbib}

\begin{document}

\title{Dual Adversarial Domain Adaptation}

\author{
Yuntao Du\institute{Nanjing University, China, duyuntao@smail.nju.edu.cn}  \and
Zhiwen Tan\institute{Nanjing University, China, yaoyueduzhen@outlook.com}   \and
Qian Chen\institute{Nanjing University, China, chenqian@smail.nju.edu.cn}   \and \\
Xiaowen Zhang\institute{Nanjing University, China, zhangxw@smail.nju.edu.cn}   \and
Yirong Yao\institute{Nanjing University, China, ytxmailg@gmail.com}   \and
Chongjun Wang\institute{Nanjing University, China, chjwang@nju.edu.cn}
}

\maketitle
\bibliographystyle{ecai}

\begin{abstract}
Unsupervised domain adaptation aims at transferring knowledge from the labeled source domain to the unlabeled target domain. Previous adversarial domain adaptation methods mostly adopt the discriminator with binary or $K$-dimensional output  to perform marginal or conditional alignment independently. Recent experiments have shown that when the discriminator is provided with domain information in both domains and label information in the source domain, it is able to preserve the complex multimodal information and high semantic information in both domains. Following this idea, we adopt a discriminator with $2K$-dimensional output to perform both domain-level and class-level alignments simultaneously in a single discriminator. However, a single discriminator can not capture all the useful information across domains and the relationships between the examples and the decision boundary are rarely explored before. Inspired by multi-view learning and latest advances in domain adaptation, besides the adversarial process between the discriminator and the feature extractor, we also design a novel mechanism to make two discriminators pit against each other, so that they can provide diverse information for each other and avoid generating target features outside the support of the source domain.  To the best of our knowledge, it is the first time to explore a dual adversarial strategy in domain adaptation. Moreover, we also use the semi-supervised learning regularization to make the representations more discriminative. Comprehensive experiments on two real-world datasets verify that our method outperforms several state-of-the-art domain adaptation methods.
\end{abstract}

\section{Introduction}

Deep neural network has achieved remarkable success in many applications \cite{b1}. However, it requires a large amount of labeled data to train the model for a good generalization. Collecting and annotating sufficient data is very expensive and time-consuming. It is a natural idea to utilize annotated data from a similar domain to help improve the performance, which is the goal of transfer learning. Generally, transfer learning aims at leveraging knowledge from a labeled source domain to use in the unlabeled target domain\cite{b2}. \textbf{{\emph{Domain adaptation}}} is a sub-filed of transfer learning, in which the feature space and label space in the source domain and target domain are the same, but the data distribution is different\cite{b2}.

It is crucial for domain adaptation to reduce the distribution discrepancy across domains\cite{b4,b5}.
Recently, many adversarial domain adaptation methods inspired by generative adversarial network\cite{b12} have been proposed. Generally, there exists a two-player game between the domain discriminator and the feature extractor\cite{b18,b33,b24,b37}. The domain discriminator is trained to distinguish the source domain from the target domain, while the feature extractor is trained to learn domain-invariant representations to confuse the discriminator. These methods all focus on domain-level(marginal) distribution alignment, while they can not promise the class-level(conditional) distribution alignment. Then, methods for class-level alignment are proposed either by using separate class-wise domain discriminators, where each discriminator is responsible for only one class\cite{b13}, or by using a discriminator with $K$-dimensional output, where $K$ is the number of class\cite{b14}.  Despite the exciting performance, the domain-level and class-level alignments are performed independently so the semantic information behind them can not be shared.

\setlength{\abovecaptionskip}{-0.4cm}  
\setlength{\belowcaptionskip}{-0.1cm}   

\begin{figure*}[htbp]
\begin{center}
\includegraphics[width = 0.85\linewidth]{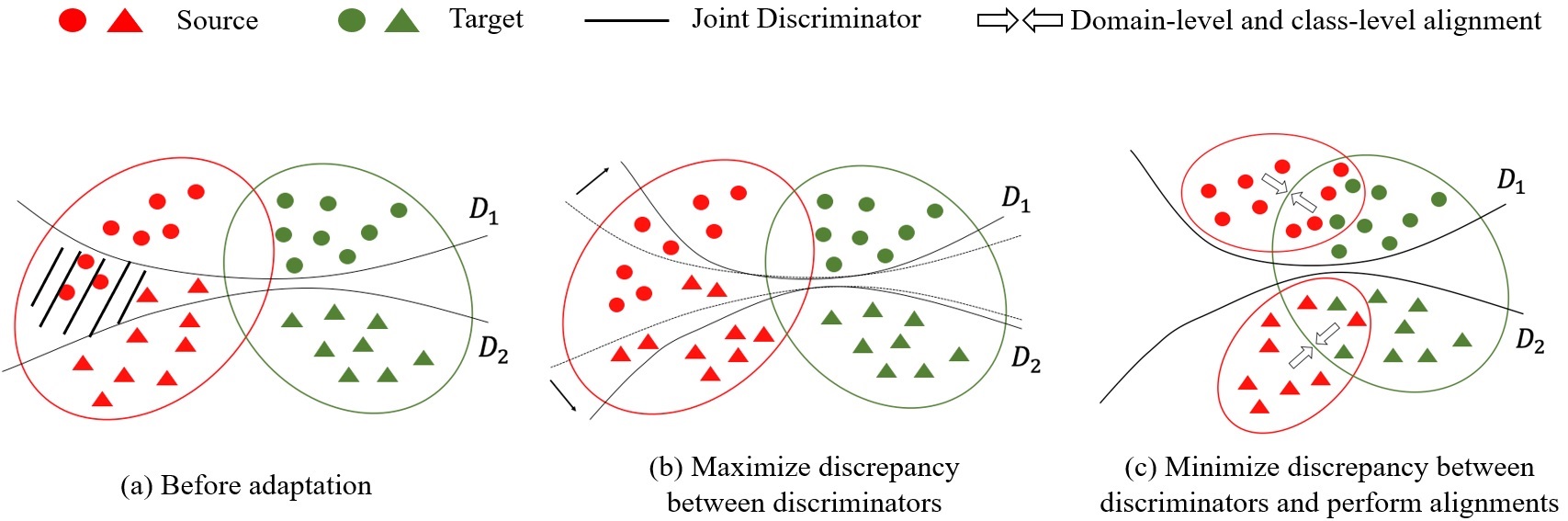}
\end{center}
\caption{An example of two discriminators with an overview of the proposed method. Note that we use a discriminators with $2K$-dimensional output, which can not only distinguish the domain label but also can classify the training data into a certain class. Discrepancy between discriminators refers to the disagreement between the prediction of two discriminators. In (a), we can see that the target samples outside the support of the source can be measured by two different discriminators. In (b), maximizing the discrepancy between the discriminators allows the two discriminators to capture different useful information, so that they can better detect the samples excluded by the support of the source. In (c), we train the feature extractor to minimize the discrepancy which can avoid generating target features outside the support of the source domain. Besides, domain-level and class-level alignments are performed to reduce the distribution discrepancy across domains. Best viewed in color.}
\label{overall}
\end{figure*}

Moreover, some methods have explored both domain-level and class-level alignments in a single discriminator. Recent experiments have shown that the informative discriminator that accesses the domain information in both domains and class information in the source domain is able to preserve the complex multimodal information and high semantic information in both domains\cite{b14,b15,b16,b17}. Instead of using a traditional discriminator with binary or $K$-dimensional output, many complex discriminators are designed. For example, a method using a discriminator with $2K$-dimensional output is proposed in \cite{b17}, where the first $K$-dimensional outputs are the known source classes, the last $K$-dimension outputs are the unknown target classes. This discriminator can distinguish the domain and class information of the training data simultaneously.

However, most existing methods adopt a single discriminator for distribution alignment, it is impossible to capture all the useful information to explore complex structure in the feature and label spaces. Besides, the previous methods do not consider the relationship between the examples and the decision boundary.  As is described in MCD\cite{b24} and shown in Figure 1(a), the samples existing far from the support of the source domain do not have discriminative features because they are not clearly categorized into certain classes.

Inspired by multi-view learning and the latest advances in domain adaptation\cite{b19,b24,b31}, we design a mechanism to make two discriminators pit against each other to solve the above issues. On the one hand, the generating complementary information can help them to capture complex semantic information. On the other hand, two adversarial discriminators(classifiers) can obtain the features, in which the support of the target is included in that of the source. Then, in order to measure the similarity between the discriminators as well as to detect the ambiguous examples, we propose to utilize the disagreement of the two discriminators on the prediction for both domains. As shown in figure 1(b), we train the two discriminators to maximize the discrepancy which can detect the target samples excluded by the support of the source.  As shown in Figure 1(c),  we train the feature extractor to minimize the discrepancy which can avoid generating target features outside the support of the source domain.

\begin{figure*}[htbp]
\begin{center}
\includegraphics[width = 0.75\linewidth]{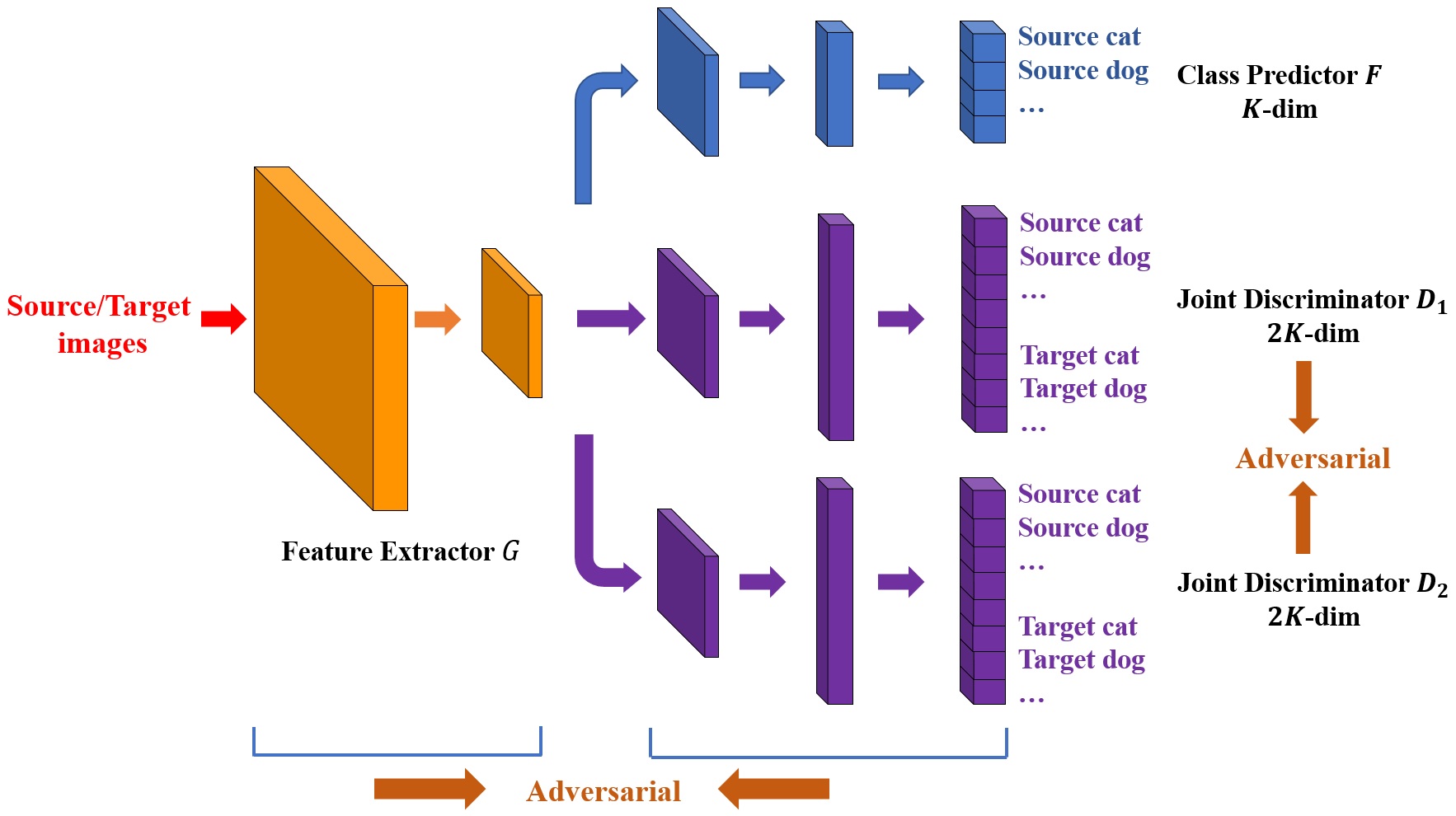}
\end{center}
\caption{ Structure of DADA algorithm. Each joint discriminator distinguishes the domain and the class of the train data to perform both domain-level and class-level alignments simultaneously in a single discriminator, while the feature extractor learns domain-invariant representations to confuse the discriminator. Two  joint discriminators are trained to pit against each other, so that they can provide diverse information for each other. Specially, there are dual adversarial processes in our algorithm. The class predictor is used to classify source examples as well as predict {\emph{pseudo labels}} for the target data.}
\label{network}
\end{figure*}

In this paper,  we focus on \textbf{{\emph{unsupervised domain adaptation}}}, where the labeled source data and unlabeled target data are available. Following this line of work, we propose a method called {{\emph{Dual Adversarial Domain Adaptation}}} (\textbf{DADA}).  As shown in Figure \ref{network}, the proposed DADA consists of a feature extractor, a class predictor and two discriminators with $2K$-dimensional output. Each discriminator learns a distribution over domain and class variables in an adversarial way, which can perform both domain-level and class-level adaptation simultaneously in a single discriminator. Besides, inspired by multi-view learning and the latest advances in domain adaptation\cite{b19,b24,b31}, we design a mechanism to make these two discriminators pit against each other, which can not only help them to learn complex semantic information from each other, but also help them to obtain the features, in which the support of the target is included in that of the source.
Note that there is a dual adversarial process in DADA, which is, to the best of our knowledge, the {\emph{first time}} to explore a dual adversarial strategy in domain adaptation. Since labels for the target data are not known, the class predictor is adopted to predict the {\emph{pseudo labels}} for the target data. Furthermore, we use the semi-supervised learning(SSL) regularization to make the extracted features more discriminative. Comprehensive experiments on two real-world image datasets are conducted and the results verify the effectiveness of our proposed method.

Briefly, our contributions lie in three folds:

(1) The discriminator with $2K$-dimensional output is adopted to perform both domain-level and class-level alignments simultaneously in a single discriminator. Moreover, the SSL regularization is used to make the representations more discriminative.

(2) We design a novel mechanism to make two discriminators pit against each other, which can make them provide diverse information for each other and avoid generating the target features outside the support of the source domain. To the best of our knowledge, it is the first time to perform a dual adversarial strategy in domain adaptation.

(3) We conduct extensive experiments on two real-world datasets  and the results validate the effectiveness of our proposed method.

\section{Related Work}

Unsupervised domain adaptation is a sub-field of transfer learning, where there are abundant labeled data in the source domain and some unlabeled data in the target domain.  Early studies focus on shallow (traditional) domain adaptation. Recently, more and more works pay attention to deep domain adaptation and adversarial domain adaptation.

\textbf{Shallow domain adaptation} The most common strategy in shallow learning is {\emph{distribution alignment}}. The distribution discrepancy between domains includes marginal distribution discrepancy and conditional distribution discrepancy. TCA\cite{b4} tries to align marginal distribution between domains, which learns a domain-invariant representation during feature mapping. Based on TCA, JDA\cite{b5} tries to algin marginal distribution and conditional distribution simultaneously. Considering the balance between marginal distribution and conditional distribution discrepancy, BDA\cite{b7} proposes a balance factor to leverage the importance of different distributions. MEDA\cite{b22} is able to dynamically evaluate the balance factor and has achieved promising performance.

\textbf{Deep domain adaptation} Most deep domain adaptation methods are based on discrepancy measure. DDC\cite{b29} embeds a domain adaptation layer into the network and minimizes Maximum Mean Discrepancy(MMD) between features of this layer. DAN\cite{b9} minimizes the feature discrepancy between the last three layers and  the mutil-kernel MMD is used to better approximate the discrepancy. Other measures are also adopted such as Kullback-Leibler (KL) divergence\cite{b26}, Correlation Alignment (CORAL)\cite{b25} and Central Moment Discrepancy (CMD)\cite{b27}. These methods can utilize the deep neural network to extract more transferable features and also have achieved promising performance.

\textbf{Adversarial domain adaptation}
Recently, adversarial learning is widely used in domain adaptation. DANN\cite{b18} use a discriminator to distinguish the source data from the target data, while the the generator learns domain-invariant feature to confuse the discriminator. Based on the theory in \cite{b3}, when maximizing the error of discriminator, it is actually approximating the $H$-distance, and minimizing the error of discriminator is actually minimizing the discrepancy between domains. ADDA\cite{b33} designs a symmetrical structure where two feature extractors are adopted. Different from DANN, MCD\cite{b24} proposes a method to minimize the $H \Delta H$-distance between domains in an adversarial way. \cite{b37} proposes a new theory using margin loss for mutli-class problem in adaptation, and based on this theory, MDD is proposed to minimize the {\emph{disparity discrepancy}} between domains.

\section{Method}

\subsection{Problem Setting}
In unsupervised domain adaptation, we are given a source domain $D_s = \{(x_i,y_i)\}_{i=1}^{n_s}$ of $n_s$ labeled source examples and a target domain $D_t = \{(x_i)\}_{i=1}^{n_t}$ of $n_t$ unlabeled target examples. The source data are drawn from the distribution $P(x_s,y_s)$ and the target data are drawn from the distribution $Q(x_t,y_t)$. Note that the i.i.d. assumption is violated, where $P(x_s,y_s) \neq Q(x_t,y_t)$. Both distributions are defined on $X \times Y$, where $Y=\{1,2,...,K\}$.
  The samples are drawn from marginal distribution(domain-level), where $x_s \sim p_s$ and $x_t \sim q_t$. Our goal is to design a deep network $x\mapsto f(x)$ to reduce the distribution discrepancy across domains in order that the generalization error $\epsilon_t(f)$ in the target domain can be bounded by source risk $\epsilon_s(f)$ plus
   the distribution discrepancy across domains\cite{b3}, where
\begin{align}
& \epsilon_t(f) = E_{(x_t,y_t) \sim Q} [f(x_t) \neq y_t] \\
& \epsilon_s(f) = E_{(x_s,y_s)\sim P} [f(x_s)\neq y_s]
\end{align}

\subsection{Overall}

As is shown in Figure \ref{network}, DADA consists of a feature extractor $G$, a class predictor $F$ and
 two joint discriminators $D_1, D_2$. Note that the output of the class predictor is $K$-dimensional
  while the output of the joint discriminator is $2K$-dimensional. Each joint discriminator pits against the feature extractor with a $2K$-way adversarial loss to learn a distribution over domain and classes variables, which can perform both domain-level and class-level alignments simultaneously in a single discriminator.
  The feature extractor $G$ aims to learn the domain-invariant representations to confuse the two joint discriminators $D_1, D_2$ so that the domain discrepancy can be reduced. We also design a mechanism to make the two joint discriminators pit against each other so that they can benefit from complementary information. Two joint discriminators are trained to increase the discrepancy between the discriminators while the feature extractor helps to perform this adversarial process by minimizing the discrepancy between the discriminators. During the adversarial process, ambiguous target samples can be detected  be pushed in the support of the source domain(section 3.5). Since the labels for the target data are unknown, it is impossible to perform class-level alignment directly. We use a class predictor $F$ trained in the source domain to predict the {\emph{pseudo labels}} for the target domain. To make the representations more discriminative and the {\emph{pseudo labels}} more accurate, we introduce semi-supervised learning  regularization, which uses the entropy minimization and {\emph{Virtual Adversarial Training}} (VAT)(section 3.6).

\subsection{Class Predictor Loss}

The class predictor $F$ is trained to classify the source  samples correctly. During the training process, it is also used to predict {\emph{pseudo labels}} for the target domain. The output of the class predictor can be written as,
\begin{equation}
f(x) = F(G(x)) \in R^K
\end{equation}

We train the network to minimize the cross entropy loss. The source classification loss of class the predictor is as follows:
\begin{align}
& \ell_{sc}(F) = E_{(x_s,y_s) \sim P}l_{CE}(f(x),y) \\
& \ell_{CE}(f(x),y) = -\left<y,\log{f(x)}\right>
\end{align}
where, the cross-entropy loss is calculated with one-hot ground-truth labels $y \in \{0,1\}^K$ and label estimates $f(x)$.
\subsection{Single Discriminator Loss}
As described in \cite{b17}, the joint discriminator is trained by a $2K$-way adversarial loss. The first $K$ are the known source classes, and the second $K$ are the unknown target classes. Such a component can learn a distribution over domain and class variables, so it can perform both domain-level and class-level alignments in a single component. The output of the joint discriminator, taking $D_1$ as an example, can be written as,
\begin{equation}
f_{D_1}(x) = D_1(G(x)) \in R^{2K}
\end{equation}

For the labeled source examples, we also train the two joint discriminators with the same classification loss . The source classification loss of the joint discriminator is defined as,
\begin{equation}
\ell_{dsc}(D_1) = E_{(x_s,y_s)\sim P} l_{CE}({D_1}(G(x_s)), [y_s,\boldsymbol{0}])
\end{equation}
where $\boldsymbol{0}$ is the zero vector of size K, chosen to make the last K joint probabilities zero for the source samples.

Similarly, to capture the label information in the target domain, we also train the discriminators using the target examples. Since the labels for the target data are not known, we use {\emph{pseudo labels}} instead. For a target example $x_{t}$, its predicted label according to the class predictor is \footnote {We use the notation x[k] for indexing the value at the kth index of the vector x}
\begin{align}
\hat y = \arg \max_{k} f(x_t)[k]
\end{align}
The  target classification loss of the joint discriminator is,
\begin{align}
\ell_{dtc}(D_1) = E_{x_t \sim q_t} l_{CE}({D_1}(x_t), [\boldsymbol{0}, \hat y_t])
\end{align}
Here, it is assumed that the source-only model can achieve reasonable performance in the target domain. In experiments, where the source-only model has poor performance initially, we use this loss after training the class predictor for a period of time.

The feature extractor $G$ is designed to confuse the joint discriminators as in DANN\cite{b18}. The basis idea is that the feature extractor can confuse the joint discriminator with the domain information, but keep label information unchanged. For example, for a source example $x_s$ with label $y_s$, the correct output of the joint discriminator should be $[y_s, \boldsymbol{0}]$, while the feature extractor fools the joint discriminator to classify it in the target domain but also using the label $y_s$, which is $[\boldsymbol{0}, y_s]$ formally.

Formally, the source alignment loss  of the joint discriminator is,
\begin{align}
\ell_{dsa1}(G) = E_{(x_s,y_s) \sim P}l_{CE}({D_1}(G(x_s)), [\boldsymbol{0},y_s])
\end{align}

Similarly, the target alignment loss of joint discriminator is defined by changing the {\emph{pseudo-label}} from $[\boldsymbol{0}, \hat y_t]$ to $[\hat y_t, \boldsymbol{0}]$,
\begin{align}
\ell_{dta1}(G) = E_{x_t \sim q_t} l_{CE}({D_1}(G(x_t)), [\hat y_t, \boldsymbol{0}])
\end{align}
The last two losses are minimized only by the feature extractor $G$. The same adversarial process is also applied in joint discriminator $D_2$.

\setlength{\abovecaptionskip}{-0.1cm}  

\subsection{Adversarial Loss Between Discriminators}

\begin{figure*}[htbp]
\centering
\subfigure[Step 2]{
\includegraphics[width=0.46\linewidth]{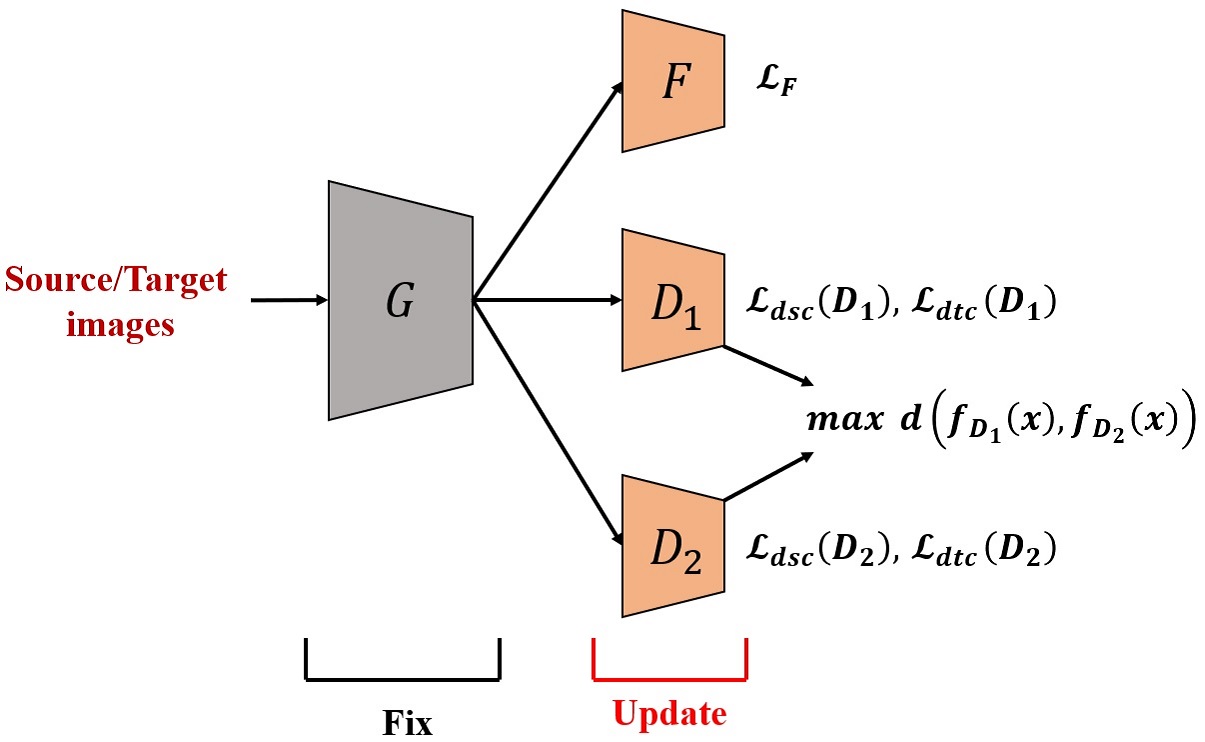}
}
\subfigure[Step 3]{
\label{fig:subfig:a}
\includegraphics[width=0.46\linewidth]{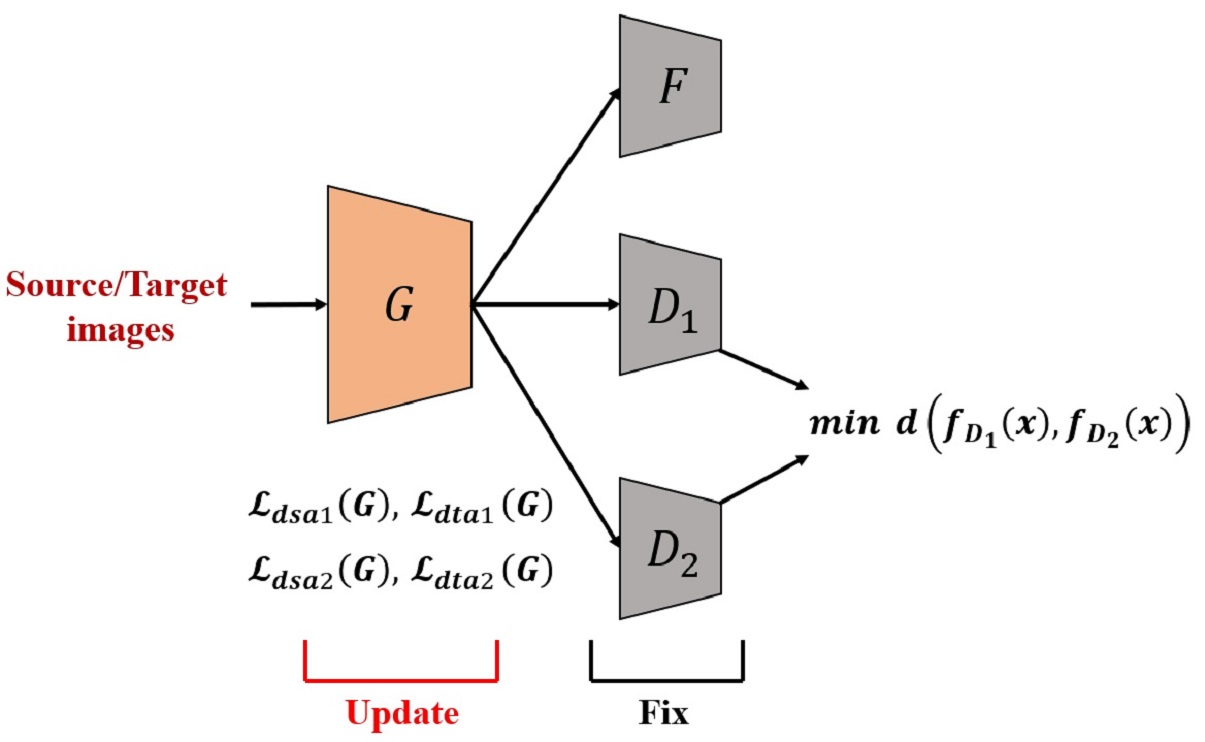}
}
\caption{Adversarial training steps of our method. There are three steps in total, step 2 and step 3 are shown in this figure. In step 2, the class predictor and two discriminators minimize the classification loss. Besides, the two discriminators pit against each other to increase the discrepancy between discriminators. In step 3, the feature extractor learns to minimize the discrepancy between discriminators as well as to confuse the discriminator in both domain and class level.}
\label{train}
\end{figure*}

Inspired by multi-view learning\cite{b19} and the latest advances in domain adatation\cite{b24,b31}, we  find that two or more similar but different components can produce diverse information so that these components can learn form each other to improve performance. In this paper, we design a mechanism to make these two discriminators pit against each other so that they can benefit from generating complementary information. Moreover, two adversarial discriminators(classifiers) can obtain the features, in which the support of the target is included in that of the source. In order to measure the similarity between the discriminators, we propose to utilize the disagreement of the two discriminators on the prediction for both domains.
The discrepancy between the two joint discriminators is defined by utilizing the absolute values of the difference between the probabilistic output as discrepancy loss:
\begin{align}
d(f_{D_1}(x),f_{D_2}(x)) = \frac{1}{K}\sum_{k=1}^K|{f_{D_1}(x)}[k] - {f_{D_2}(x)}[k]|
\end{align}

We firstly train the discriminators to increase their discrepancy. It can not only help different discriminators to capture different information, but also detect the target
samples excluded by the support of the source\cite{b24}. The objective is as follows,
\begin{equation}
\begin{aligned}
& \max_{D_1,D_2} \ell_{d} \\
\end{aligned}
\end{equation}
\begin{equation}
\begin{aligned}
\ell_{d}= &E_{x_s \sim p_s}[d(f_{D_1}(x_s),f_{D_2}(x_s))]     \\
                   + &E_{x_t \sim q_t}[d(f_{D_1}(x_t),f_{D_2}(x_t))]
\end{aligned}
\end{equation}

Moreover, the feature extractor is trained to minimize the discrepancy for fixed discriminators. On the one hand, minimizing the discrepancy can make these two joint discriminators not too far away from each other, thus making them similar. On the other hand, minimizing the discrepancy can avoid generating target features outside the support of the source domain\cite{b24}. The objective is as follows,
\begin{align}
\min_{G} \ell_{d}
\end{align}

\subsection{SSL Regularization Loss}
After the distribution alignment, the discrepancy across domains can be smaller. In this case, we can approximate the unsupervised domain adaptation as a semi-supervised learning problem. On this bias, many previous works have explored semi-supervised learning(SSL)  regularization in  domain adaptation\cite{b17,b30} and made sufficient improvements. Although lacking of labels, a large amount of unlabeled data can be used to bias the classifier boundaries to pass through the regions containing low density data. Thus, the learned representation can become more discriminative. Entropy minimization is a widely used  regularization method to achieve this goal. In our method, the class predictor is also trained to minimize the target entropy loss, which is defined as follows,
\begin{align}
& \ell_{te}(F) = E_{(x_t,y_t)\sim q}\ell_E(f(x))
\end{align}
where $\ell_E(f(x)) = -\sum_kf(x)[k] \cdot \log{f(x)[k]}$. However, minimizing entropy is only applicable to locally-Lipschitz classifiers\cite{b21}. So we propose to explicitly incorporate the locally-Lipschitz condition via {\emph{virtual adversarial training}}(VAT) and add the following losses to the objective,
\begin{align}
& \ell_{svat}(F) = E_{x_s \sim p_s}[\max_{||r|| \leq \epsilon}\ell_{CE}(f_i(x_s)||f_i(x_s+r))] \\
& \ell_{tvat}(F) = E_{x_t \sim q_t}[\max_{||r|| \leq \epsilon}\ell_{CE}(f_i(x_t)||f_i(x_t+r))]
\end{align}

\subsection{Overall Objective}

We combine the objective functions discussed in section3.4-3.6 and divide our training produce into three steps.

\textbf{Step 1} We only use the source data to train the feature extractor $G$, the class predictor $F$ as well as the joint discriminators $D_1,D_2$.  We minimize the source classification loss of the class predictor and joint discriminators. After the classifier and joint discriminators are trained to classify the source samples correctly, we will go on the next step. The objective of this step is shown as follows,
\begin{align}
\min_{G,F,D_1,D_2} \ell_{sc}(F) + \lambda_{dsc1}\ell_{dsc}(D_1) + \lambda_{dsc2}\ell_{dsc}(D_2)
\end{align}

\textbf{Step 2} We fix the feature extractor, and update the class predictor and the discriminators. We use both the source and target data to update the model. This process corresponds to Step 2 in Figure \ref{train}. We have three sub-objectives. The first one is to minimize the source and target classification loss of the joint discriminators. The second one is to minimize the source classification of the class predictor as well as  the SSL regularization loss. Without this loss, we experimentally found that the performance dropped. The last one is to increase the discrepancy between the discriminators. The objective of this step is shown as follows,
\begin{align}
&\min_{F,D_1,D_2}  \ell_F+ \ell_{D_1} + \ell_{D_2} - \lambda_{d} \ell_{d} \\
&\ell_{D_1} =  \lambda_{dsc1}\ell_{dsc}(D_1) +  \lambda_{dtc1}\ell_{dtc}(D_1) \\
&\ell_{D_2} =  \lambda_{dsc2}\ell_{dsc}(D_2) +  \lambda_{dtc2}\ell_{dtc}(D_2)
\end{align}
\begin{equation}
\begin{aligned}
\ell_F = \ell_{sc}(F) + \lambda_{svat}\ell_{svat}(F) + \lambda_{te}\ell_{te}(F) \\ + \lambda_{tvat}\ell_{tvat}(F)
\end{aligned}
\end{equation}
\textbf{Step 3} We fix the class predictor and the discriminators, and update the feature extractor. We train the model by minimizing the source and target alignment loss of joint discriminators as well as the discrepancy between discriminators. The objective of this step is shown as follows,
\begin{equation}
\begin{aligned}
\min_{G}  \lambda_{dsa1}\ell_{dsa1}(G) + \lambda_{dta1}\ell_{dta1}(G) + \lambda_{dsa2}\ell_{dsa2}(G) + \\
\lambda_{dta2}\ell_{dta2}(G) + \lambda_{d}\ell_{d}
\end{aligned}
\end{equation}
Step2 and Step3 are repeated alternately in our method. We are concerned on that the feature extractor , class predictor and joint discriminators are trained in an adversarial manner so that they can classify the source samples correctly as well as promote the cross-domain discrepancy decreasing.

\begin{table*}[!htbp]
\centering
\caption{Classification accuracy (\%) on Office-31 for unsupervised domain adaptation with ResNet-50.}
\begin{tabular}{p{3.0cm}p{1.4cm}<{\centering}p{1.4cm}<{\centering}p{1.4cm}<{\centering}p{1.4cm}<{\centering}p{1.4cm}<{\centering}p{1.4cm}<{\centering}p{1.4cm}<{\centering}}
\toprule
Method&   A$\rightarrow$W&   D$\rightarrow$W&   W$\rightarrow$D&   A$\rightarrow$D&   D$\rightarrow$A&   W$\rightarrow$A&  Avg\\
\midrule
ResNet-50\cite{b36}&  68.4$\pm$0.2&   96.7$\pm$0.1&   99.3$\pm$0.1&   68.9$\pm$0.2&   62.5$\pm$0.3&   60.7$\pm$0.3&   76.1\\
DAN\cite{b9}&  80.5$\pm$0.4&   97.1$\pm$0.2&   99.6$\pm$0.1&   78.6$\pm$0.2&   63.6$\pm$0.3&   62.8$\pm$0.2&   80.4\\
DANN\cite{b18}& 82.6$\pm$0.4&   96.9$\pm$0.2&   99.3$\pm$0.2&   81.5$\pm$0.4&   68.4$\pm$0.5&   67.5$\pm$0.5&   82.7\\
ADDA\cite{b33}& 86.2$\pm$0.5&   96.2$\pm$0.3&   98.4$\pm$0.3&   77.8$\pm$0.3&   69.5$\pm$0.4&   68.9$\pm$0.5&   82.9\\
MADA\cite{b13}& 90.0$\pm$0.1&   97.4$\pm$0.1&   99.6$\pm$0.1&   87.8$\pm$0.2&   70.3$\pm$0.3&   66.4$\pm$0.3&   85.2\\
VADA\cite{b40}& 86.5$\pm$0.5&   98.2$\pm$0.4&   99.7$\pm$0.2&   86.7$\pm$0.4&   70.1$\pm$0.4&   70.5$\pm$0.4&   85.4\\
GTA\cite{b39}&  89.5$\pm$0.5&   97.9$\pm$0.3&   99.7$\pm$0.2&   87.7$\pm$0.5&   \textbf{72.8}$\pm$0.3&   71.4$\pm$0.4&   86.5\\
MCD\cite{b24}&  88.6$\pm$0.2&   98.5$\pm$0.1&   100.0$\pm$.0&   92.2$\pm$0.2&   69.5$\pm$0.1&   69.7$\pm$0.3&   86.5\\
RCA\cite{b17}&  93.8$\pm$0.2&   98.4$\pm$0.1&   100.0$\pm$.0&   91.6$\pm$0.2&   68.0$\pm$0.2&   70.2$\pm$0.2&   87.0\\
CDAN\cite{b34}& 93.1$\pm$0.1&   98.6$\pm$0.1&   100.0$\pm$.0&   92.9$\pm$0.2&   71.0$\pm$0.3&   69.3$\pm$0.3&   87.5\\
\midrule
DADA(without SSL)& 94.0$\pm$0.2&  98.4$\pm$0.2&  100.0$\pm$.0&   91.6$\pm$0.2&  68.3$\pm$0.1&   69.7$\pm$0.2&   87.0 \\
DADA&        \textbf{94.5}$\pm$0.2&   \textbf{98.7}$\pm$0.2&   \textbf{100.0}$\pm$.0&   \textbf{93.6}$\pm$0.3&   69.5$\pm$0.3&   \textbf{71.5}$\pm$0.2&   \textbf{88.0}\\
\bottomrule
\end{tabular}
\label{res_1}
\end{table*}

\begin{table*}[!htbp]
\centering
\caption{Classification accuracy (\%) on ImageCLEF-DA for unsupervised domain adaptation with ResNet-50.}
\begin{tabular}{p{3.0cm}p{1.4cm}<{\centering}p{1.4cm}<{\centering}p{1.4cm}<{\centering}p{1.4cm}<{\centering}p{1.4cm}<{\centering}p{1.4cm}<{\centering}p{1.4cm}<{\centering}}
\toprule
Method&   I$\rightarrow$P&   P$\rightarrow$I&   I$\rightarrow$C&   C$\rightarrow$I&   C$\rightarrow$P&   P$\rightarrow$C&  Avg\\
\midrule
ResNet-50\cite{b36}&  74.8$\pm$0.3&   83.9$\pm$0.1&   91.5$\pm$0.3&   78.0$\pm$0.2&   65.5$\pm$0.3&   91.2$\pm$0.3&   80.7\\
DAN\cite{b9}&  74.5$\pm$0.4&   82.2$\pm$0.2&   92.8$\pm$0.2&   86.3$\pm$0.4&   69.2$\pm$0.4&   89.8$\pm$0.4&   82.5\\
DANN\cite{b18}& 75.0$\pm$0.3&   86.0$\pm$0.3&   96.2$\pm$0.4&   87.0$\pm$0.5&   74.3$\pm$0.5&   91.5$\pm$0.6&   85.0\\
MADA\cite{b13}& 75.0$\pm$0.3&   87.9$\pm$0.2&   96.0$\pm$0.3&   88.8$\pm$0.3&   75.2$\pm$0.2&   92.2$\pm$0.3&   85.8\\
CDAN\cite{b34}& 76.7$\pm$0.3&   90.6$\pm$0.3&   97.0$\pm$0.4&   90.5$\pm$0.4&   74.5$\pm$0.3&   93.5$\pm$0.4&   87.1\\
TAT\cite{b35}&  78.8$\pm$0.2&   92.0$\pm$0.2&   97.5$\pm$0.3&   92.0$\pm$0.3&   \textbf{78.2}$\pm$0.4&   94.7$\pm$0.4&   88.9\\
RCA\cite{b17}&  78.7$\pm$0.2&   92.8$\pm$0.2&   97.7$\pm$0.3&   92.0$\pm$0.2&   77.0$\pm$0.3&   \textbf{95.0}$\pm$0.3&   88.9\\
DADA&      \textbf{79.0}$\pm$0.3& \textbf{93.2}$\pm$0.2& \textbf{98.2}$\pm$0.2&   \textbf{92.3}$\pm$0.2&   77.8$\pm$0.3&   \textbf{95.0}$\pm$0.3&   \textbf{89.3}\\
\bottomrule
\end{tabular}
\label{res_2}
\end{table*}

\subsection{Theoretical Understanding}

Most existing domain adaptation methods are based on the domain adaptation theory proposed in \cite{b3}. The generalization error in the target domain $\epsilon_{t}(h)$ is bounded by three terms: (1) the expected error $\epsilon_s(h)$ in the source domain, (2) the $H\Delta H$-distance $d_{H\Delta H}(S,T)$ between domains, measuring the disagreement of two hypothesis $h,h' \in H \Delta H$ and (3) the error $\lambda$ of the ideal hypothesis $h^*$ in both domains. The main Theorem in \cite{b3} is shown as below,

\newtheorem{thm}{\bf Theorem}
\begin{thm}
Let $H$ be the hypothesis space, then for any $h$ in the hypothesis space,
\begin{align}
\epsilon_t(h) \leq \epsilon_s(h) + \frac{1}{2}d_{H \Delta H}(X_s,X_t) + \lambda
\end{align}
where
\begin{align*}
d_{H \Delta H}(D_s,D_t) = 2 \sup _{h,h' \in H}|Pr_{x \sim X_s}[h(x) \neq h'(x)] \\
- Pr_{x \sim X_t}[h(x) \neq h'(x)] | \\
h^* =  \arg \min_{h \in H} \epsilon_s(h) + \epsilon_t(h),  \lambda = \epsilon_s(h^*) + \epsilon_t(h^*)
\end{align*}
\end{thm}
 The binary class discriminator  is used in DANN to distinguish where the train data come from(source or target), if it is not able to classify the data correctly, the cross-domain discrepancy can be reduced. In fact, it makes the $d_{H \Delta H}(S,T)$ become small. However, an example in \cite{b31} shows that even the $d_{H \Delta H}(S,T)$ is zero, there also exists discrepancy across domains. Because if the conditional distribution between domains is not matched, the third term $\lambda$ can be large. In our method, the joint discriminators can align the conditional distribution across domains, so $\lambda$ will be small. Besides, two adversarial discriminators can provide diverse information including label formation for each other, so it is helpful to match the conditional distribution better. Moreover, predicting {\emph{pseudo labels}} for the target data may cause some mistakes, while the SSL  regularization can help correct some errors, thus also making it better to match the conditional distribution across domains.

\section{Experiments}

We evaluate the proposed method with many state-of-the-art domain adaptation methods on two image datasets. Codes will be available at  \url{https://github.com/yaoyueduzhen/DADA}.

\subsection{Setup}
\textbf{Office-31}\cite{b32} is the most widely used dataset for domain adaptation, with 4,652 images and 31 categories collected from three distinct domains: Amazon (\textbf{A}), Webcam (\textbf{W}) and DSLR (\textbf{D}). From this dataset, we build six transfer tasks: \textbf{A} $\rightarrow$ \textbf{W}, \textbf{D} $\rightarrow$ \textbf{W}, \textbf{W} $\rightarrow$ \textbf{D}, \textbf{A} $\rightarrow$ \textbf{D}, \textbf{D} $\rightarrow$ \textbf{A}, and \textbf{W} $\rightarrow$ \textbf{A}.

\textbf{ImageCLEF-DA}\cite{b10} is a dataset organized by selecting the 12 common classes shared by three public datasets (domains): Caltech-256 (\textbf{C}), ImageNet ILSVRC 2012 (\textbf{I}), and Pascal VOC 2012 (\textbf{P}). We evaluate all methods on six transfer tasks: \textbf{I} $\rightarrow$ \textbf{P}, \textbf{P} $\rightarrow$ \textbf{I}, \textbf{I} $\rightarrow$ \textbf{C}, \textbf{C} $\rightarrow$ \textbf{I}, \textbf{C} $\rightarrow$ \textbf{P}, and \textbf{P} $\rightarrow$ \textbf{C}.

We compare Dual Adversarial Domain Adaptation (\textbf{DADA}) with several state-of-the-art domain adaptation methods: Deep Adaptation Network (\textbf{DAN})\cite{b9}, Domain Adversarial Neural Network (\textbf{DANN})\cite{b18}, Adversarial Discriminative Domain Adaptation (\textbf{ADDA})\cite{b33}, Multi-Adversarial Domain Adaptation (\textbf{MADA})\cite{b13}, Virtual Adversarial Domain Adaptation (\textbf{VADA})\cite{b40}, Generate to Adapt(\textbf{GTA})\cite{b39}, Maximum Classifier Discrepancy (\textbf{MCD})\cite{b24}, Conditional Domain Adversarial Network (\textbf{CDAN})\cite{b34}, Transferable Adversarial Training (\textbf{TAT})\cite{b35} and Regularized Conditional Alignment (\textbf{RCA})\cite{b17}.

\subsection{Implementation Details}

Following the standard protocols for unsupervised domain adaptation\cite{b18,b10}, all labeled source samples and unlabeled target samples participate in the training stage. We compare the average classification accuracy based on five random experiments. The results of other methods are reported in the corresponding papers except RCA which is reimplemented by ourselves.

We use PyTorch to implement our method and use \textbf{ResNet-50}\cite{b36} pretrained on ImageNet\cite{b38} as the feature extractor. The class predictor and two discriminators are both two-layer fully connected networks with a width of 1024. We train these new layers and feature extractor using back-propagation, where the learning rates of these new layers are 10 times that of the feature extractor. We adopt mini-batch SGD with the momentum of 0.9 and use the same learning rate annealing strategy as \cite{b18}: the learning rate is adjusted by $\eta_p = \eta_0(1+\alpha p)^{-\beta}$, where $p$ is the training progress changing from 0 to 1, and $\eta_0 = 0.04, \alpha = 10, \beta = 0.75$.

We fix $\lambda_{d} = 1.0$ and search the rest hyperparameters over $\lambda_{dsc1},\lambda_{dsc2} \in \{0.1,0.5,1.0\}$, $\lambda_{dtc1},\lambda_{dtc2} \in \{0.1,1.0,10.0\}$, $\lambda_{svat},\lambda_{tvat} \in \{0.0,0.1,1.0\}$, $\lambda_{te} \in \{0.1,1.0\}, \lambda_{dsa1},\lambda_{dta1},\lambda_{dsa2},\lambda_{dta2} \in \{0.1,1.0\}$. We also search for the upper bound of the adversarial perturbation in VAT, where $\epsilon \in \{0.5,1.0,2.0\}$. The optimal hyperparameters on Office-31 dataset are shown in Table \ref{param}.

\begin{table}
\centering
\caption{Optimal hyperparameters on Office-31 dataset.}
\begin{tabular}{ccccccc}
\toprule
Tasks&   A$\rightarrow$W&   D$\rightarrow$W&   W$\rightarrow$D&   A$\rightarrow$D&   D$\rightarrow$A&   W$\rightarrow$A \\
\midrule
$\lambda_{dsc1}$&   1.0& 1.0& 1.0& 1.0& 0.5& 1.0\\
$\lambda_{dsc2}$&   1.0& 1.0& 1.0& 1.0& 0.5& 1.0\\
$\lambda_{dtc1}$&   1.0& 1.0& 1.0& 1.0& 1.0& 1.0\\
$\lambda_{dtc2}$&   1.0& 1.0& 1.0& 1.0& 1.0& 1.0\\
$\lambda_{d}$&      1.0& 1.0& 1.0& 1.0& 1.0& 1.0\\
$\lambda_{svat}$&   1.0& 1.0& 1.0& 1.0& 0.0& 1.0\\
$\lambda_{tvat}$&   1.0& 1.0& 1.0& 1.0& 1.0& 1.0\\
$\lambda_{te}$&     0.1& 0.1& 0.1& 0.1& 1.0& 1.0\\
$\lambda_{dsa1}$&   0.1& 0.1& 0.1& 0.1& 0.1& 0.1\\
$\lambda_{dta1}$&   0.1& 0.1& 0.1& 0.1& 0.1& 0.1\\
$\lambda_{dsa2}$&   0.1& 0.1& 0.1& 0.1& 0.1& 0.1\\
$\lambda_{dta2}$&   0.1& 0.1& 0.1& 0.1& 0.1& 0.1\\
$\epsilon$&         0.5& 0.5& 0.5& 0.5& 0.5& 0.5\\
\bottomrule
\end{tabular}
\label{param}
\end{table}

\subsection{Results}
The results on Office-31 dataset are shown in Table \ref{res_1}. As we can see, our method outperforms baseline methods in most tasks and achieves the best result with average accuracy. Compared with DANN and ADDA, which only perform domain-level alignments using a binary class discriminator, our method performs not only domain-level but also class-level alignments and outperforms them. MADA considers domain-level and class-level alignment, but it constrains each discriminator to be responsible for only one class. Our method avoids this limitation by adopting  $2K$-dimensional discriminators where the classes can share information. The $2K$-dimensional joint discriminator is also used in RCA, but we train two discriminators in an adversarial manner so that they can provide  complementary information for each other. Moreover, we clearly observe that our method can also perform well on D$\rightarrow$A and  W$\rightarrow$A with relatively large domain shift and imbalanced domain scales.

The results on ImageCLEF-DA are shown in Table \ref{res_2}. We have several findings based on the results. Firstly, all methods are better than ResNet-50, which is a source-only model trained without exploiting the target data in a standard supervised learning setting. Our method increases the accuracy from 80.7\% to 89.3\%. Secondly, the above comparisons with baseline methods on Office-31 are also the same on ImageCLEF-DA, which verifies the effectiveness of our method. Thirdly, DADA outperforms the baseline methods on most transfer tasks, but with less room for improvement. This is reasonable since the three domains in ImageCLEF-DA are of equal size and balanced in each category, which makes domain adaptation easy.

\subsection{Analysis}

\textbf{Ablation Study.} We study the effect of SSL  regularization by removing entropy minimization and VAT losses from our method($\lambda_{te} = \lambda_{svat} = \lambda_{tvat} = 0$), which is denoted by DADA(without  SSL). The results on Office-31 dataset are reported in Table \ref{res_1}.  Results show that without SSL regularization, our method can perform  better in two tasks(\textbf{A}$\rightarrow$\textbf{W}, \textbf{W}$\rightarrow$\textbf{D}) than baseline methods , but the average accuracy of all tasks is decreased by 1.0\% compared to the proposed DADA. The results validate the effectiveness of SSL regularization.

\begin{figure}[htbp]
\centering
\subfigure[Source Only]{
\includegraphics[width=0.47\linewidth]{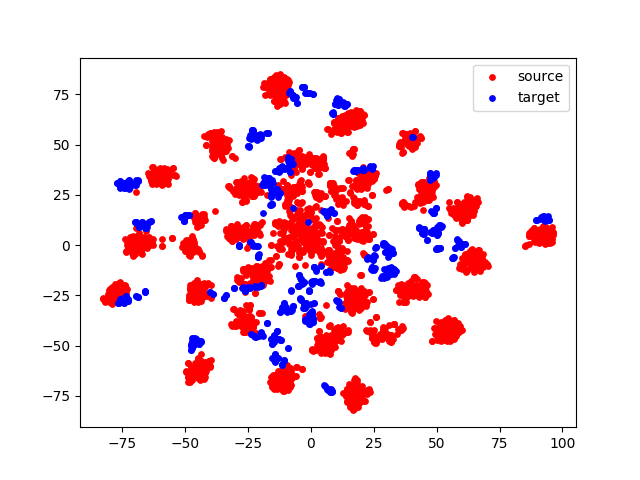}
}
\subfigure[Adapted(Ours)]{
\label{fig:subfig:a}
\includegraphics[width=0.47\linewidth]{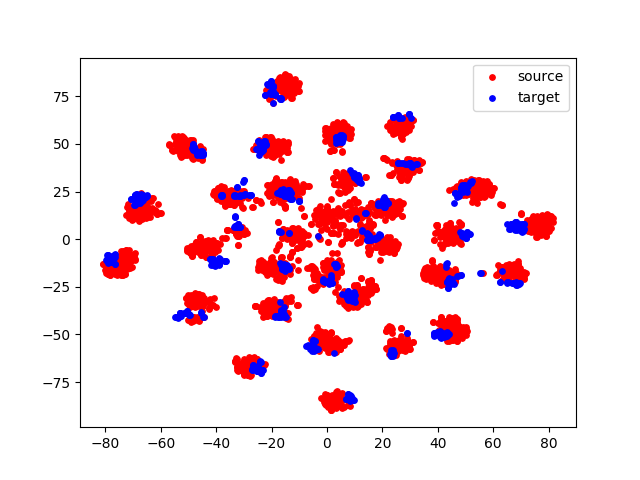}
}
\caption{Visualization of features obtained from the feature extractor of task \textbf{A}$\rightarrow$\textbf{W} using t-SNE\cite{b41}.Red and blue points indicate the source and target samples respectively. We can see that applying our method makes the target samples more discriminative.}
\label{Visualization}
\end{figure}

\textbf{Feature Visualization.} In Figure \ref{Visualization}, we visualize the feature representations of task \textbf{A}$\rightarrow$\textbf{W}(31 classes) by t-SNE\cite{b41} using the source-only method and DADA. The source-only method is trained without exploiting the target training data in a standard supervised learning setting using the same learning procedure. As we can see, source and target samples are better aligned for DADA than the source-only method. This shows the advance of our method in discriminative prediction.

\setlength{\abovecaptionskip}{-0.1cm}  

\textbf{Distribution Discrepancy.} The $A$-distance is a measure of distribution discrepancy, defined as $dist_A = 2 (1-2\epsilon)$, where $\epsilon$ is the test error of a classifier trained to distinguish the source from the target. We use $A$-distance as a measure of the transferability of feature representations. Table \ref{A-distance} shows the cross-domain $A$-distance for tasks \textbf{A}$\rightarrow$\textbf{W}, \textbf{W}$\rightarrow$\textbf{D}. We compute the $A$-distance of our method based on the output of the feature extractor G, which turns out to be the smallest of all methods.

\begin{table}
\centering
\caption{Cross-domain $A$-distance of different approaches.}
\begin{tabular}{lp{2.0cm}<{\centering}p{2.0cm}<{\centering}}
\toprule
Method&   D$\rightarrow$W&   W$\rightarrow$A\\
\midrule
ResNet-50\cite{b36}&  1.27&   1.86\\
DANN\cite{b18}&       1.23&   1.44\\
MCD\cite{b24}&        1.22&   1.60\\
DADA&     \textbf{1.14}& \textbf{1.18}\\
\bottomrule
\end{tabular}
\label{A-distance}
\end{table}

\section{Conclusion}
In this paper, we propose a method called {\emph{Dual Adversarial Domain Adaptation}}(\textbf{DADA}), which is able to performe both domain-level and class-level alignments simultaneously in a single discriminator.  Besides, inspired by mutil-view learning, we design a  novel mechanism to make  two discriminators pit against each other, and encourage them to learn  different semantic information to benefit from each other.  Moreover, SSL regularization is used to make the representations more discriminative so that the predicted {\emph{pseudo labels}} can be more accurate.  We conduct comprehensive experiments and the results verify the effectiveness of our proposed method.

\bibliography{ecai}
\end{document}